\documentclass[10pt,conference]{IEEEtran}
\IEEEoverridecommandlockouts
\usepackage{cite}
\usepackage{float}
\usepackage{amsmath,amssymb,amsfonts}
\usepackage{graphicx, enumitem}
\usepackage[caption=false]{subfig}
\usepackage{textcomp}
\usepackage[ruled,vlined,linesnumbered]{algorithm2e}
\usepackage[table,dvipsnames]{xcolor}
\usepackage{tikz}
\usepackage{bm}
\usepackage{color,soul}
\usepackage{makecell}
\usepackage{booktabs}
\usepackage{multirow}
\usepackage{url}
\usepackage{tikz}
\usepackage{listings}
\usepackage{xcolor}
\usepackage{orcidlink}
\usepackage[none]{hyphenat}
\SetKwIF{If}{ElseIf}{Else}{if}{}{else if}{else}{end if}%
 
\newcolumntype{L}[1]{>{\raggedright\let\newline\\\arraybackslash\hspace{0pt}}m{#1}}
\newcolumntype{C}[1]{>{\centering\let\newline\\\arraybackslash\hspace{0pt}}m{#1}}
\newcolumntype{R}[1]{>{\raggedleft\let\newline\\\arraybackslash\hspace{0pt}}m{#1}} 

\def\BibTeX{{\rm B\kern-.05em{\sc i\kern-.025em b}\kern-.08em
    T\kern-.1667em\lower.7ex\hbox{E}\kern-.125emX}}

\expandafter\def\expandafter\normalsize\expandafter{%
    \normalsize%
    \setlength\abovedisplayskip{4pt}%
    \setlength\belowdisplayskip{4pt}%
}

\begin{document}

\title{Optimizing Drilling Trajectories in Multi‑Material Structures Using Quantum Annealers
	\thanks{This work was supported by the Basque Government through ELKARTEK Program under Grants KK-2025/00074 (Newhegaz project).}
}
\author{
	\IEEEauthorblockN{Eneko Osaba\IEEEauthorrefmark{2}\IEEEauthorrefmark{1}\orcidlink{0000-0001-7863-9910}, and
        Antonio Rubio-Mateos\IEEEauthorrefmark{2}\orcidlink{0000-0002-3701-003X}}
	\IEEEauthorblockA{\IEEEauthorrefmark{2}TECNALIA, Basque Research and Technology Alliance (BRTA), 48160 Derio, Bizkaia, Spain}
	\IEEEauthorblockA{\IEEEauthorrefmark{1}Corresponding author. Email: eneko.osaba@tecnalia.com}}
    
\maketitle

\IEEEpubidadjcol	

\begin{abstract}
In this work we present a quantum‑hybrid optimization approach for planning drilling trajectories on multi‑material structures, with the objective of minimizing total energy consumption. The proposed formulation models the drilling process as a constrained routing problem in which each hole must be visited while accounting for heterogeneous material properties, material‑dependent drilling costs and displacement energy, as well as tool‑wear penalties. Our method incorporates realistic industrial constraints, including open‑path routing and mandatory grouping of holes belonging to the same material to avoid frequent tool‑bit changes. The modeled problem is solved using D‑Wave’s Leap Constrained Quadratic Model Hybrid Solver. To evaluate performance, we benchmark instances ranging from 10 to 55 holes with varying material distributions and cost matrices. The results highlight the capability of quantum‑hybrid optimization to handle complex multi‑physics constraints and to generate energy‑efficient drilling sequences under realistic manufacturing conditions.
\end{abstract}


\section{Introduction}\label{sec:intro}

Modern manufacturing increasingly relies on multi-material structures whose heterogeneous properties make drilling trajectory planning a challenging optimization task. Suboptimal hole sequencing and frequent tool-bit changes increase energy consumption, machining time, and tool wear, particularly in complex parts with diverse materials \cite{kim2016effect}. As a result, sectors such as aerospace, automotive, and precision machining require trajectory planners that jointly optimize sequencing and manufacturability constraints \cite{khodabakhshi2023minimization}.

Although classical optimization techniques perform well in many routing-related problems, incorporating multi-physics interactions and heterogeneous constraints remains challenging at scale \cite{dewil2019critical}. In this context, hybrid quantum-classical computing has emerged as a promising alternative, combining classical optimization with quantum resources to tackle complex constrained problems \cite{callison2022hybrid}. Recent studies have reported encouraging results in manufacturing applications such as scheduling, production planning, and routing. However, most quantum-assisted approaches remain focused on Traveling Salesman Problem (TSP)‑like formulations and do not address the specific requirements of drilling on multi-material structures \cite{osaba2022systematic}. This gap motivates the formulation proposed in this work, which explicitly captures these industrial constraints within a constrained optimization framework.

This paper proposes a hybrid quantum-classical approach for energy‑efficient drilling trajectory planning on multi‑material structures. We model the process as a constrained routing problem using a Constrained Quadratic Model (CQM), which integrates: \textit{i)} material‑grouping constraints to minimize tool‑bit changes, \textit{ii)} displacement‑related energy terms reflecting the physical motion of the drilling head, \textit{iii)} material‑dependent drilling costs and \textit{iv)} tool‑wear penalties. To address this problem, we employ D‑Wave’s LeapCQMHybrid solver \cite{dwave_cqm_whitepaper}, which orchestrates classical search with quantum‑annealing cycles. We empirically benchmark instances ranging from 10 to 55 holes with heterogeneous material layouts and cost matrices, showing that the hybrid pipeline can generate energy‑aware, manufacturable drilling sequences under realistic constraints.

From an industrial standpoint, the proposed system enables material‑aware, energy‑conscious trajectories that remain manufacturable under realistic constraints. From a methodological standpoint, keeping constraints explicit in the CQM avoids ill‑conditioning due to large penalty coefficients and yields a more stable, interpretable, and scalable modeling pathway for multi‑physics drilling operations.

The remainder of this paper is organized as follows: Section \ref{sec:back} provides the necessary background to contextualize
our work. Section \ref{sec:problem} formalizes the problem and details the CQM. Sections \ref{sec:setup} and  \ref{sec:analysis} describe the solver workflow, experimental setup, and results. Finally, Section \ref{sec:conc} is dedicated to conclusions and future work.


\section{Background}\label{sec:back}

Trajectory planning and sequencing problems appear across a wide range of industrial and autonomous-systems applications, from aerospace trajectory optimization to surface inspection, multi-robot coordination, and logistics \cite{mir2022survey}. Classical optimization continue to deliver strong performance in many routing-like scenarios \cite{liu2023heuristics}. However, their scalability can be challenged when problems involve multi-physics constraints, heterogeneous costs, or complex operational rules, motivating the exploration of alternative paradigms \cite{meraliyev2025comprehensive}.

In parallel, quantum computing, particularly quantum annealing and variational quantum algorithms, has emerged as a complementary tool for combinatorial routing and path-planning tasks \cite{cattelan2024modeling}. Quantum annealing has been applied to trajectory optimization through QUBO formulations \cite{carbone2023cutting}, while hybrid quantum-classical approaches have shown competitive performance in coverage path planning and related applications in robotics and aerospace \cite{rao2024quantum}. Despite the limitations of current NISQ-era hardware, these studies highlight the growing potential of quantum-assisted optimization for routing and sequencing problems \cite{abbas2024challenges}. Beyond robotics and aerospace, quantum-enabled optimization has also been explored in industrial and logistics applications \cite{curuliuc2026qubo,osaba2024exploring}. Furthermore, recent hybrid quantum-classical studies have effectively addressed real‑world trajectory‑planning constraints, such as time windows, vehicle-capacity limits, simultaneous pickup-and-delivery, heterogeneous fleets, incomplete graphs, open routes, multi-robot coordination, and airspace separation \cite{osaba2025optimizing,osaba2025quantum,rao2024quantum,haba2025routing}. 

Nevertheless, to the best of our knowledge, no existing quantum approach addresses drilling-trajectory optimization on multi-material structures. This problem involves material-dependent drilling and displacement costs, explicit tool-wear modelling, mandatory grouping of same-material holes to minimize tool-bit changes, and open-path manufacturability requirements that are not captured in previous quantum applications \cite{xu2022review}. This gap motivates a domain‑specific hybrid quantum-classical system that encodes realistic drilling constraints. Accordingly, we formulate the drilling‑trajectory optimization as a constrained routing problem via a CQM, which integrates directly with the D‑Wave LeapCQMHybrid solver used in this work.

\section{Mathematical Formulation}\label{sec:problem}

In this section, we present in detail the mathematical formulation of the Drill Routing Problem (DRP) addressed in this research. The problem is defined over a set of drilling points $N=\{0,1,\dots,n-1\}$, each associated with a material $m(i) \in M$, where $M$ denotes the set of all materials involved in the drilling task. The objective is to determine a feasible drilling sequence that satisfies the material block constraints while minimizing the total traversal cost.

\subsection{Decision Variables}

Two types of variables are used to define the problem. The first type is associated with the encoding employed to represent the drilling trajectory. In this work, a node-based binary encoding is used, which is frequently adopted in studies of quantum computing applied to routing problems \cite{feld2019hybrid}. Thus, for each node $i \in N$ and time step $t \in T = \{0,\dots,n-1\}$, a decision variable $x_{i,t}$ is defined, taking value 1 if node $i$ is drilled at time $t$, and 0 otherwise.

Two further sets of variables are also defined, grouped in a category called \textit{material block variables}, which are introduced to enforce the material grouping required to ensure solution feasibility. This way, for each material $m \in M$ and $t \in T$, two different variables are defined:

\begin{itemize}
    \item $\mathrm{start}_{m,t}$, which is 1 if a block of material $m$ starts at time $t$, and 0 otherwise.
    \item $\mathrm{inblock}_{m,t}$, which is 1 if in time $t$ the drill is inside a block of material $m$, and 0 otherwise.
\end{itemize}

\subsection{Objective Function}

The transition cost between two drilling points depends on the material that will be drilled next. This modelling choice is grounded in the physical characteristics of the drilling process: different materials require different drill bits, each with specific mechanical properties such as hardness, coating, geometry, and weight. Consequently, the energy required to move and stabilize the drill head before machining the next hole varies depending on the bit to be used. For example, titanium (Ti, grade Ti‑6Al‑4V) typically requires a heavier and more robust drill bit than aluminium alloy (AA, grade AA7075‑T6), leading to higher energy consumption during tool displacement. This difference reflects an underlying material‑dependent physical behaviour; therefore, the traversal cost in the proposed model is defined as a function of the material corresponding to the destination node.

Throughout the paper, \(c^{m}_{i,j}\) denotes the displacement energy required to move from hole \(i\) to hole \(j\) when drilling material \(m \in M\). For notational convenience, we define \(c_{i,j} \equiv c^{m(j)}_{i,j}\), i.e., the traversal cost depends on the material of the destination hole. Without loss of generality, and for the sake of clarity, we assume that two materials are involved in the drilling process: AA and Ti. Thus, let the cost of transitioning from node $i$ to node $j$ depend on the material of the destination node:
\begin{equation}
c_{i,j} =
\begin{cases}
c^{AA}_{i,j}, & \text{if } m(j)=AA, \\[2mm]
c^{Ti}_{i,j}, & \text{if } m(j)=Ti.
\end{cases}
\end{equation}

This formulation is directly generalizable to any number of materials $m \in M$ by defining a family of material-specific cost matrices $c^{m}_{i,j}$. In the objective function, the instantiated traversal cost is always given by $c^{m(j)}_{i,j}$, i.e., the cost associated with moving from node $i$ to node $j$ depends on the material of the destination node $j$. Furthermore, all energy-related quantities in the objective function (namely, the displacement energy $c^{m(j)}_{i,j}$, the drilling energy $d_{m(i)}$, and the tool-wear penalties $p_t$) are expressed in kilowatt-hours (kWh). So, the objective is to minimize the overall route cost:

\begin{equation}
\begin{aligned}
\min \Bigg(
& \sum_{\substack{i,j \in N \\ i \neq j}} \sum_{t=0}^{n-2}
c^{\,m(j)}_{i,j}\, x_{i,t}\, x_{j,t+1}
& + \sum_{i=0}^{n-1} \sum_{t=0}^{n-1}
d_{m(i)}\, x_{i,t} \\
& + \sum_{t=0}^{n-1}
p_t \sum_{i=0}^{n-1} x_{i,t}
\Bigg)
\end{aligned}
\tag{2}
\end{equation}

This function is composed of three complementary energy terms. The first term models the displacement-related energy required to move the drilling head between consecutive holes, where the traversal cost \(c^{m(j)}_{i,j}\) depends on the material of the destination hole. This formulation captures the effect of material-specific tool characteristics on the energy required for motion.

The second term explicitly models the drilling energy, where \(d_{m(i)}\) denotes the energy required to drill a hole made of material
\(m(i)\). This term captures the material-dependent machining effort, reflecting differences in hardness, cutting resistance, and  required spindle load across materials.

The third term represents tool-wear-related overconsumption, where \(p_t\) models the additional energy required at time step \(t\) due to progressive tool degradation along the drilling sequence. This formulation allows the model to capture cumulative wear effects without explicitly tracking tool state variables.

\subsection{Constraints}

First, three different constraints are needed for ensuring the consistency of the route:

    \begin{itemize}
        \item \textit{Location-consistency}: Only one node can be drilled at each time step.
        \begin{equation}
        \sum_{i=0}^{n-1} x_{i,t} = 1,
        \qquad \forall t \in T
        \end{equation}
        \item \textit{Drilling-consistency}: Each node is drilled exactly once.
        \begin{equation}
        \sum_{t=0}^{n-1} x_{i,t} = 1,
        \qquad \forall i \in N
        \end{equation}
        \item \textit{Origin first}: The origin, defined as the coordinate (0,0) and indexed by \(o\), must be visited at time 0.
        \begin{equation}
        x_{o,0} = 1
        \end{equation}
        
    \end{itemize}

Furthermore, for a trajectory to be feasible, the material block consistency constraints must also be satisfied. To ensure this, a set of constraints has been introduced. Specifically, four groups of constraints are defined:

\begin{itemize}
     \item \textit{Block continuation}: If a block of material $m$ starts at time $t$, then the next $\min(L,\,n-t)$ time steps must remain within that block:
    
    \begin{equation}
    \begin{aligned}
        \mathrm{inblock}_{m,k} &\ge \mathrm{start}_{m,t}, \\
        &\forall m \in M,\; \forall t \in T, \\
        &\forall k \in [t,\,\min(t+L-1,\,n-1)]
    \end{aligned}
    \end{equation}
    where $L$ denotes the minimum block length that must be respected.
    
    \item \textit{Material consistency inside a block}: a block of material $m$ can only be active if a node of material $m$ is drilled.
    \begin{equation}
    \sum_{i : m(i)=m} x_{i,t} \ge \mathrm{inblock}_{m,t},
    \qquad \forall m \in M,\; \forall t \in T
    \end{equation}
    \item \textit{Block starts upon material entry}: This constraint ensures that a block for material $m$ starts exactly when the trajectory first drills a node of that material. In other words, whenever the material at time step $t$ is $m$ and it was not $m$ at time $t-1$, the variable $\mathrm{start}_{m,t}$ must activate. So, for all $m \in M$ and $t>0$:
    \begin{equation}
    \mathrm{start}_{m,t}
    \ge
    \left(
    \sum_{i : m(i)=m} x_{i,t}
    -
    \sum_{i : m(i)=m} x_{i,t-1}
    \right)
    \end{equation}
    For $t=0$, the block simply starts if the first drilled node is made of material $m$.
    \begin{equation}
    \mathrm{start}_{m,0}
    \ge
    \sum_{i : m(i)=m} x_{i,0}
    \end{equation}
\end{itemize}

This formulation supports any number of materials $|M| \ge 2$ without requiring structural changes.


\section{Experimental Setup}
\label{sec:setup}

This section presents the experimental evaluation of the proposed quantum-classical approach introduced in this paper. 

\subsection{Quantum solver details} \label{sec:solver}

Going deeper into the rationale for adopting the CQM formulation, this kind of model generalizes the traditional QUBO approach by allowing equality and inequality constraints to be declared explicitly in the model, thereby avoiding the manual embedding of constraints through problem‑dependent penalty weights. In practice, each constraint is internally transformed into an energy‑contributing term through an automatic penalty‑scaling mechanism, which balances feasibility enforcement and objective minimization without introducing numerical instability. This largely mitigates the need for empirical penalty tuning, which is a well‑known source of ill‑conditioning and convergence issues in pure QUBO formulations, where linear or mixed‑integer constraints must be encoded using large penalty coefficients. In our implementation, all decision variables are binary or integer, and both the objective function and constraints are expressed using the \texttt{dimod.ConstrainedQuadraticModel()} interface, after which the complete model is compiled and submitted to the hybrid solver.

\begin{figure}[t]
 \centering
 \includegraphics[width=0.65\linewidth]{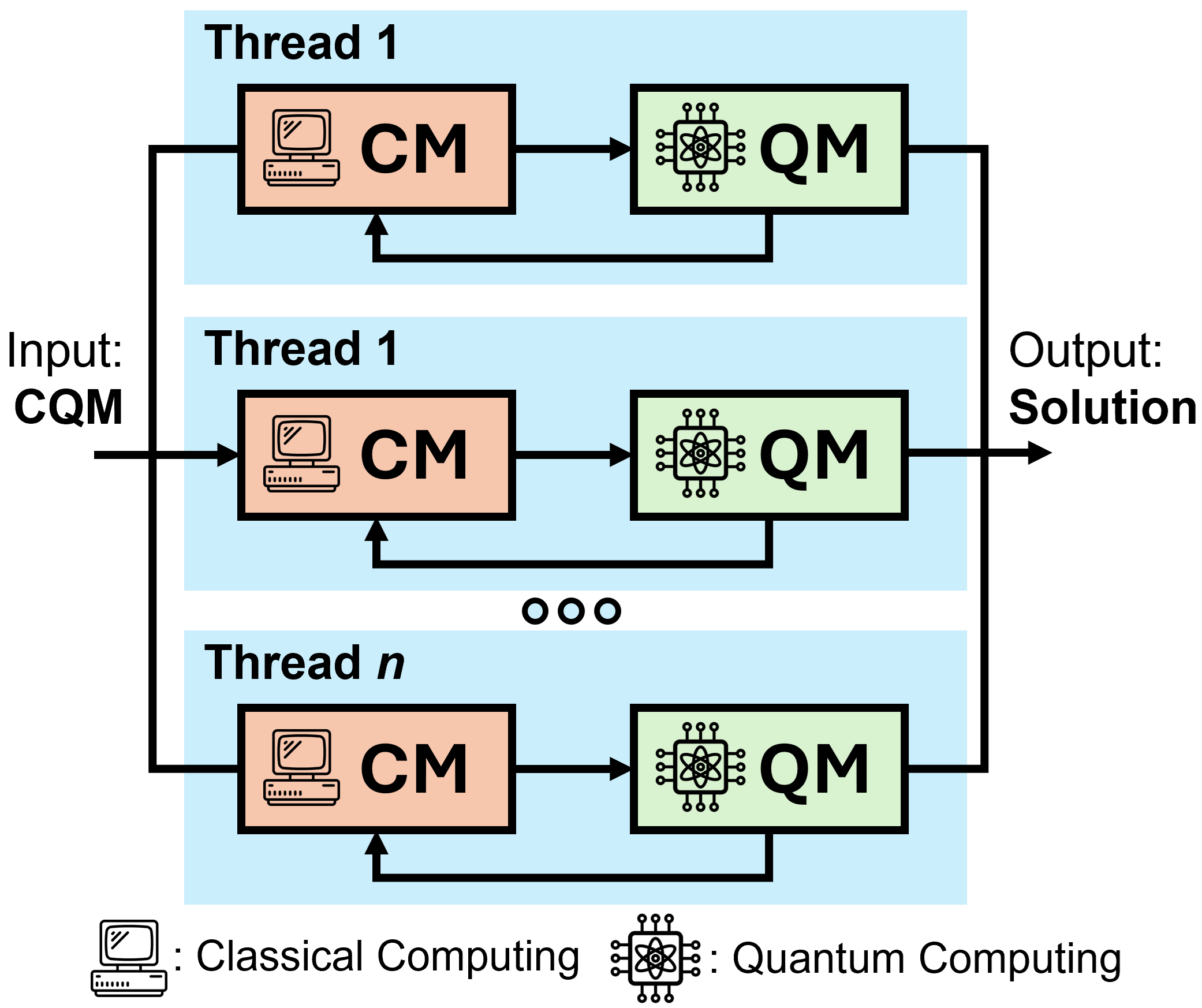}
 \caption{D-Wave’s CQM scheme. CM = Classical Module. QM = Quantum Module.}
 \label{fig:cqm}
\end{figure}

Regarding the employed solver, the LeapCQMHybrid integrates classical and quantum resources within an asynchronous, multi-branch pipeline, which is graphically represented in Figure \ref{fig:cqm}. Each submitted instance is decomposed into parallel branches comprising two tightly coupled modules: \textit{i}) a \emph{Classical Module} (CM), responsible for large-scale exploration using classical metaheuristics (e.g., local search, tabu search, simulated annealing), candidate construction, and subproblem reduction; and \textit{ii}) a \emph{Quantum Module} (QM), which performs quantum annealing cycles on \texttt{D-Wave Advantage2\_system1.11} (4596 qubits, Zephyr topology). The resulting low-energy samples are ranked and asynchronously fed back to the CM.

The CM and QM interact through an adaptive feedback mechanism in which the scheduler monitors the progress of the classical module and triggers quantum queries when stagnation is detected or additional diversification is required. These quantum invocations return diverse low-energy samples that complement the classical search, enabling the overall workflow to escape local stagnation and explore alternative regions of the solution space.  Hence, the quantum component acts as an accelerator for diversification and refinement rather than a full stand-alone solver, while the classical component ensures scalability and native handling of constraints.

Overall, this scheme allows us to: \textit{i}) preserve the original constrained structure of the DRP; \textit{ii}) avoid fragile penalty engineering; and \textit{iii}) exploit quantum-guided sampling to enhance robustness and solution quality under tight time limits.

\subsection{Benchmark description} \label{sec:benchmark}
To evaluate the performance of our DRP formulation, we generated a set of benchmark instances consisting of drilling structures with heterogeneous material distributions, varying distances between holes, and different combinations of AA and Ti. Since no standard dataset exists for the DRP, all instances were generated synthetically, following realistic industrial criteria provided by machining experts. As mentioned before, all energy values used in the benchmark instances (displacement, drilling, and wear-related terms) are provided in kWh, ensuring consistency across components of the objective.

For each structure, holes were placed on a 2D grid emulating fuselage sections or composite panels commonly found in aerospace manufacturing. Material layouts were designed to include blocks of different sizes and interleavings of AA and Ti to test the robustness of the material-block constraints. For interested readers and to support reproducibility, the complete benchmark
dataset is publicly available\footnote{\url{http://dx.doi.org/10.17632/zh27vhp3dn.1}}.

\section{Performance Analysis}
\label{sec:analysis}

\begin{table}[t]
\setlength{\tabcolsep}{2pt}
\centering
\caption{Feasibility quality and feasible-solution diversity across time limits. 
For each instance, we report: total number of variables and constraints, 
mean feasible-solution rate in percentage (FSR), mean percentage of broken constraints 
in infeasible solutions, and number of distinct feasible solutions found. Best results in bold.}
\label{tab:feasibility_quality}
\resizebox{1\columnwidth}{!}{
    \begin{tabular}{c r c|cccc|cccc|cccc|cccc}
    \toprule
    \multicolumn{3}{c|}{\textbf{Instance}} &
    \multicolumn{4}{c|}{\textbf{5s}} &
    \multicolumn{4}{c|}{\textbf{10s}} &
    \multicolumn{4}{c|}{\textbf{15s}} &
    \multicolumn{4}{c}{\textbf{20s}} \\
    \cmidrule(lr){1-3}
    \cmidrule(lr){4-7}
    \cmidrule(lr){8-11}
    \cmidrule(lr){12-15}
    \cmidrule(lr){16-19}
    \textbf{Name} & \textbf{\#Var.} & \textbf{\#Constr.} &
    \textbf{FSR} & \textbf{Brok.} & \textbf{Feas.} & &
    \textbf{FSR} & \textbf{Brok.} & \textbf{Feas.} & &
    \textbf{FSR} & \textbf{Brok.} & \textbf{Feas.} & &
    \textbf{FSR} & \textbf{Brok.} & \textbf{Feas.} & \\
    \midrule
    DRP\_10 & 140   & 143 & 78.7 & \textbf{5.81} & 26 & & 81.0 & 5.84 & 26 & & \textbf{94.0} & 9.82 & \textbf{28} & & 90.9 & 9.54 & 25 & \\
    DRP\_15 & 285   & 223 & 77.8 & 6.83 & 42 & & 78.6 & 7.25 & 40 & & 90.7 & 7.83 & 55 & & \textbf{92.4} & \textbf{6.81} & \textbf{57} & \\
    DRP\_20 & 480   & 303 & 77.7 & 5.38 & 42 & & 77.0 & 5.60 & 43 & & 89.4 & 4.52 & 63 & & \textbf{90.5} & \textbf{2.66} & \textbf{65} & \\
    DRP\_25 & 725   & 383 & 66.7 & 8.56 & 35 & & 66.8 & 7.47 & 34 & & 79.7 & 6.06 & 54 & & \textbf{81.7} & \textbf{5.16} & \textbf{55} & \\
    DRP\_30 & 1020   & 463 & 60.1 & 8.22 & 28 & & 60.5 & 7.15 & 28 & & 67.0 & \textbf{6.47} & \textbf{43} & & 66.4 & 7.88 & \textbf{44} & \\
    DRP\_35 & 1365   & 543 & 54.3 & 9.61 & 22 & & 54.9 & 9.10 & 22 & & 62.6 & 8.04 & \textbf{37} & & \textbf{62.9} & \textbf{7.25} & \textbf{37} & \\
    DRP\_40 & 1760   & 623 & 55.7 & 6.48 & 21 & & 54.3 & 8.95 & 20 & & \textbf{60.3} & \textbf{6.45} & \textbf{35} & & 60.0 & 8.89 & 34 & \\
    DRP\_45 & 2205   & 703 & 53.1 & 8.57 & 18 & & 55.9 & \textbf{8.46} & 19 & & \textbf{59.9} & 8.47 & 31 & & 59.1 & 9.62 & \textbf{32} & \\
    DRP\_50 & 2700   & 783 & 55.5 & 8.91 & 18 & & 53.3 & 9.89 & 18 & & 56.7 & \textbf{8.54} & 28 & & \textbf{57.6} & 8.58 & \textbf{30} & \\
    DRP\_55 & 3245  & 863 & 51.4 & 8.80 & 15 & & 53.6 & 9.71 & 16 & & \textbf{51.3} & \textbf{7.54} & 26 & & 56.2 & 8.76 & \textbf{29} & \\
    \bottomrule
    \end{tabular}
}
\end{table}

In this section, we report a comprehensive experimental evaluation of the proposed approach. The analysis is structured into three complementary parts: \textit{i)} the feasibility and constraint compliance of the solutions; \textit{ii)} the quality of the obtained routes; and \textit{iii)} the structural properties of the material blocks. Together, these subsections provide a complete picture of the solver’s performance under realistic settings.

\subsection{Feasibility analysis}
\label{subsec:fea_quality}

This subsection evaluates the performance of the proposed hybrid CQM-based solver on the benchmark instances described previously. Since the LeapCQMHybrid solver returns a set of candidate solutions per submission, each associated with an objective value, feasibility status, and constraint residuals, the analysis extends beyond the single best feasible solution and examines the overall quality of the complete sample set.


\paragraph{Evaluation protocol}
For each instance and for four different time limits (5, 10, 15 and 20 seconds), we executed $K=10$ independent runs (different seeds).  For every run, the complete sample set returned by the hybrid solver was processed through an external feasibility checker, which verifies all DRP constraints under a numerical tolerance of $10^{-6}$.  From this analysis, we extract three complementary metrics:

\begin{enumerate}
    \item \textit{Feasible-solution rate (FSR)}: the proportion of feasible solutions within the sample set. This metric reflects the solver’s robustness and its tendency to generate feasible trajectories beyond the single best solution.
    
    \item \textit{Mean constraint-breaches percentage in infeasible solutions}:  for candidates that fail to satisfy one or more DRP constraints, we compute the average percentage of broken constraints with respect to the total number of constraints in the instance.  This quantifies how ``close'' the non-feasible candidates lie to feasibility, providing insight into how the hybrid algorithm navigates the constraint landscape.
    
    \item \textit{Diversity of feasible solutions}:  the number of \emph{distinct} feasible solutions found across the sample set.     Distinctness is measured based on the visit sequence (i.e., the pattern of drilled nodes over time), capturing the solver's ability to explore multiple structurally valid drilling trajectories.
\end{enumerate}

These three metrics offer complementary information: while FSR captures feasibility robustness, constraints-breaches analysis measures the structural quality of the non-feasible portion of the sample set, and diversity quantifies the exploration capability of the hybrid workflow.

\paragraph{Results} Table~\ref{tab:feasibility_quality} summarizes the results. The first column reports the number of constraints in each instance, which naturally grows with problem size and material heterogeneity. For all time limits, the CQM solver consistently produces at least one feasible solution. However, the fraction of feasible candidates in the sample set increases with the computation time, with the best performance consistently observed at 15s and 20s. This behavior aligns with the internal hybrid mechanisms described in D‑Wave’s technical documentation and reported in the literature, where classical search and quantum‑guided refinement operate asynchronously, and longer solver times allow a greater number of refinement cycles to be executed. \cite{dwave_cqm_whitepaper,v2023hybrid}. 

The mean percentage of broken constraints among infeasible solutions decreases as the time limit increases, indicating that the solver progressively concentrates the search around the feasible region of the solution space. Moreover, the feasible-solution diversity also increases with time, suggesting that the hybrid method is capable not only of refining solutions but also of exploring alternative feasible trajectories, which is a desirable property in industrial settings where different drilling sequences can have similar energy profiles or different practical advantages.

Overall, the trends observed across all three metrics confirm that the proposed hybrid approach can reliably enforce the structural requirements of the DRP formulation while maintaining the exploratory benefits of hybrid classical-quantum search.  Increased time limits translate into a higher proportion of feasible candidates and enriched feasible diversity. These findings validate the suitability of CQM-based hybrid optimization for handling the constraint-rich and multi-physics nature of the DRP.

\subsection{Objective quality analysis}
\label{subsec:obj_quality}

\begin{figure}[t]
    \centering
    \subfloat[DRP\_20 (Block)]{%
        \includegraphics[width=0.4\linewidth]{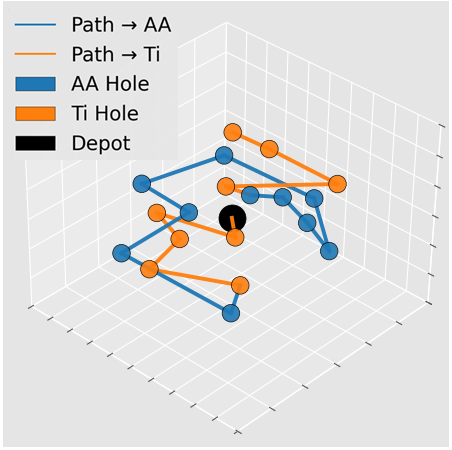}%
        \label{fig:drp20_block}%
    }
    \hspace{1cm}%
    \subfloat[DRP\_25 (Block)]{%
        \includegraphics[width=0.4\linewidth]{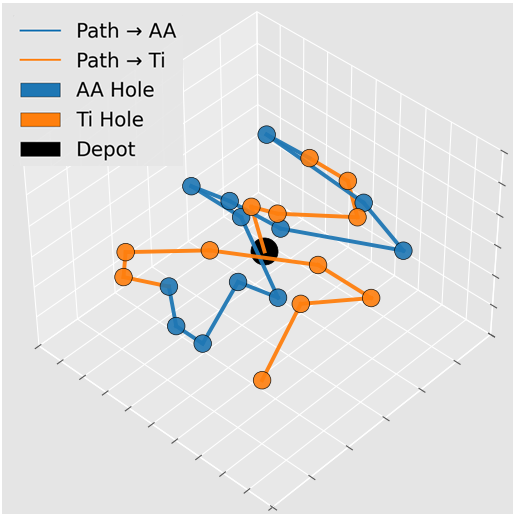}%
        \label{fig:drp25_block}%
    }

    \subfloat[DRP\_20 (No-block)]{%
        \includegraphics[width=0.4\linewidth]{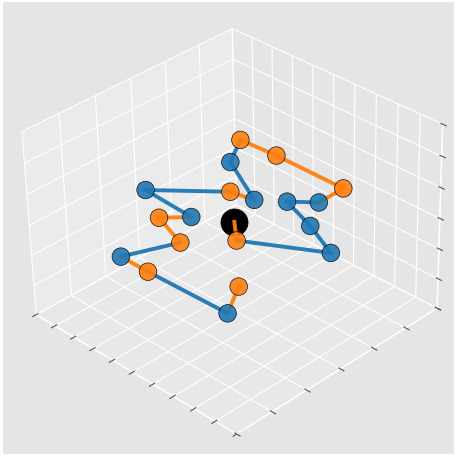}%
        \label{fig:drp20_noblock}%
    }
    \hspace{1cm}%
    \subfloat[DRP\_25 (No-block)]{%
        \includegraphics[width=0.4\linewidth]{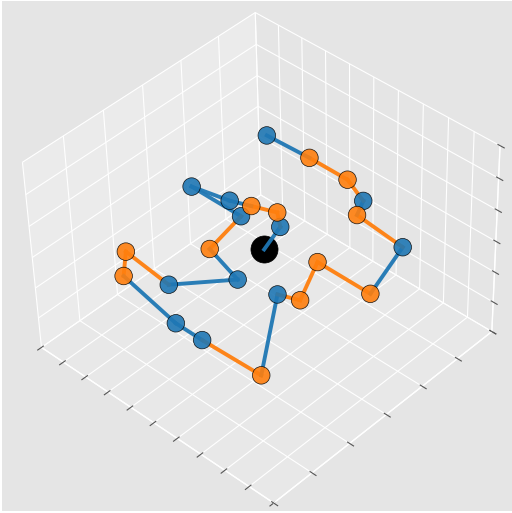}%
        \label{fig:drp25_noblock}%
    }

    \caption{Side-by-side comparison of drilling trajectories for instances \texttt{DRP\_20} and \texttt{DRP\_25} with solver time $=15$\,s. Top row shows routes satisfying material-block consistency (CQM), and bottom row shows unconstrained geometric TSP solutions.}
    \label{fig:drp_comparison_matrix}
\end{figure}

This subsection complements the feasibility study by evaluating the \emph{quality} of the solutions obtained under different time budgets. For each instance, and for time limits of $5$, $10$, $15$, and $20$ seconds, we conduct $K=10$ independent runs and report aggregate statistics across those runs. The analysis focuses on decomposing the cost into physically meaningful components and on quantifying the structural overheads induced by realistic constraints.

When analysing the numerical results, it is also instructive to visually compare the trajectories obtained when the material‑block constraint is enforced with those produced by an unconstrained geometric solver. Figure \ref{fig:drp_comparison_matrix} illustrates representative cases for two instances of different sizes. The unconstrained solutions, computed through a classical TSP baseline, minimize geometric distance only and therefore exhibit highly interleaved material patterns. In contrast, the CQM‑based solutions incorporate the block‑consistency requirements, producing trajectories that are slightly longer in geometric terms but significantly more coherent from a manufacturing standpoint. These examples highlight the practical effect of the material‑grouping constraint: although enforcing blocks increases traversal length by discouraging abrupt material alternation, it also generates trajectories that align with tooling and manufacturability principles.

With this qualitative intuition in place, we now turn to a quantitative evaluation of objective quality. The corresponding numerical results are reported in Table~\ref{tab:objective_quality_split}, where the total energy is decomposed into traversal, drilling, and tool-wear components and compared against the geometric baselines.

\begin{table}[t]
\setlength{\tabcolsep}{2.2pt}
\centering
\caption{Objective-quality summary across time limits (means over $K=10$ runs). 
For each instance, we report: total runtime, best total energy, traversal-only energy, 
traversal baseline with geometric gap (\%), and the penalty gap (total minus traversal). All energy values are reported in kWh.}
\label{tab:objective_quality_split}

\resizebox{1\columnwidth}{!}{
    \begin{tabular}{l|ccccc|ccccc}
    \toprule
    \multirow{2}{*}{\textbf{Instance}} &
    \multicolumn{5}{c|}{\textbf{5s}} &
    \multicolumn{5}{c}{\textbf{10s}} \\
    \cmidrule(lr){2-6} \cmidrule(lr){7-11}
     & \textbf{Tot.~rt.}
     & \textbf{Best~energy}
     & \textbf{Traversal}
     & \textbf{Baseline}
     & \textbf{Gap}
     & \textbf{Tot.~rt.}
     & \textbf{Best~energy}
     & \textbf{Traversal}
     & \textbf{Baseline}
     & \textbf{Gap} \\
    \midrule
    DRP\_10 & 7.25 & \textbf{22.21} & \textbf{19.71} & 15.47 & 2.50
       & 15.23 & \textbf{22.21} & \textbf{19.71} & 15.47 & 2.50 \\

    DRP\_15 & 8.05 & \textbf{26.94} & \textbf{22.98} & 19.11 & 3.96
           & 15.20 & \textbf{26.94} & \textbf{22.98} & 19.11 & 3.96 \\
    
    DRP\_20 & 11.00 & 31.56 & 26.39 & 18.30 & 5.17
           & 15.08 & 31.30 & 26.13 & 18.30 & 5.17 \\
    
    DRP\_25 & 9.18 & 39.78 & 33.10 & 20.70 & 6.68
           & 14.90 & 37.16 & 30.49 & 20.42 & 6.67 \\
    
    DRP\_30 & 7.68 & 44.88 & 36.81 & 20.13 & 8.07
           & 15.74 & 46.00 & 37.94 & 19.46 & 8.06 \\
    
    DRP\_35 & 8.12 & 52.33 & 42.57 & 26.33 & 9.76
           & 16.04 & 48.43 & 38.72 & 25.38 & 9.71 \\
    
    DRP\_40 & 8.97 & 62.68 & 51.48 & 28.88 & 11.20
           & 15.63 & 60.64 & 49.39 & 27.81 & 11.25 \\
    
    DRP\_45 & 8.55 & 71.11 & 58.07 & 31.44 & 13.04
           & 16.82 & 66.11 & 53.08 & 29.49 & 13.03 \\
    
    DRP\_50 & 10.00 & 82.38 & 67.56 & 31.22 & 14.82
           & 17.07 & 75.17 & 60.62 & 28.82 & 14.55 \\
    
    DRP\_55 & 9.56 & 90.12 & 73.28 & 33.47 & 16.84
           & 17.38 & 89.53 & 72.71 & 32.76 & 16.82 \\
    \bottomrule
    \end{tabular}
}

\resizebox{1\columnwidth}{!}{
    \begin{tabular}{l|ccccc|ccccc}
    \toprule
    \multirow{2}{*}{\textbf{Instance}} &
    \multicolumn{5}{c|}{\textbf{15s}} &
    \multicolumn{5}{c}{\textbf{20s}} \\
    \cmidrule(lr){2-6} \cmidrule(lr){7-11}
     & \textbf{Tot.~rt.} 
     & \textbf{Best~energy} 
     & \textbf{Traversal} 
     & \textbf{Baseline} 
     & \textbf{Gap}
     & \textbf{Tot.~rt.} 
     & \textbf{Best~energy} 
     & \textbf{Traversal} 
     & \textbf{Baseline} 
     & \textbf{Gap} \\
    \midrule
    DRP\_10 & 16.09 & \textbf{22.21} & \textbf{19.71} & 15.47 & 2.50
        & 24.42 & \textbf{22.21} & \textbf{19.71} & 15.47 & 2.50 \\

    DRP\_15 & 17.70 & \textbf{26.94} & \textbf{22.98} & 19.11 & 3.96
            & 24.89 & \textbf{26.94} & \textbf{22.98} & 19.11 & 3.96 \\
    
    DRP\_20 & 18.18 & \textbf{31.22} & \textbf{26.05} & 18.30 & 5.17
            & 24.57 & 31.25 & 26.08 & 18.30 & 5.17 \\
    
    DRP\_25 & 18.28 & \textbf{36.17} & \textbf{29.48} & 20.53 & 6.69
            & 25.21 & 36.51 & 29.80 & 20.41 & 6.71 \\
    
    DRP\_30 & 18.64 & 40.29 & 32.23 & 19.49 & 8.06
            & 25.03 & \textbf{40.09} & \textbf{32.01} & 19.31 & 8.08 \\
    
    DRP\_35 & 18.59 & 45.89 & 36.17 & 25.54 & 9.72
            & 25.23 & \textbf{45.61} & \textbf{35.87} & 25.11 & 9.74 \\
    
    DRP\_40 & 19.45 & 51.77 & 40.52 & 27.34 & 11.25
            & 25.70 & \textbf{51.18} & \textbf{39.93} & 27.52 & 11.25 \\
    
    DRP\_45 & 19.16 & 61.53 & 48.50 & 29.63 & 13.03
            & 26.79 & \textbf{57.11} & \textbf{44.07} & 29.05 & 13.04 \\
    
    DRP\_50 & 19.19 & 77.00 & 62.12 & 29.40 & 14.88
            & 27.08 & \textbf{72.47} & \textbf{57.59} & 28.52 & 14.88 \\
    
    DRP\_55 & 20.60 & 84.82 & 68.06 & 31.95 & 16.76
            & 27.68 & \textbf{76.61} & \textbf{59.74} & 31.51 & 16.87 \\
    \bottomrule
    \end{tabular}
}

\end{table}

\paragraph{Protocol and aggregation} Each run returns a sample set of candidate solutions from which we retain the \emph{best feasible} one (lowest total energy) for that run. From that single representative solution we extract the following quantities and then compute their mean over the $K$ runs:

\begin{enumerate}
    \item \textit{Total runtime (mean)}: end-to-end wall-clock time. Note that the this runtime always exceeds the solver time because, in addition to the time spent inside the CQM solver, the workflow includes several model‑building and pre‑processing steps. These extra processes account for the difference between the solver time (which governs the 5, 10, 15 and 20s experimental settings) and the larger end‑to‑end runtime observed in practice.
    \item \textit{Best total energy (mean)}: objective value of the best feasible solution (traversal + drilling energy + tool‑wear penalties); 
    \item \textit{Traversal (mean)}: the displacement-related term of the best feasible solution (excluding drilling energy and tool‑wear penalties). 
\end{enumerate}

\paragraph{Geometric baseline for traversal} To contextualize the efficiency of the displacement component, we compute a traversal-only baseline using the same LeapCQMHybrid solver, but through separate and dedicated executions. Specifically, for each instance, the CQM formulation is re-executed while retaining only the displacement-related terms in the analysis, thereby disregarding drilling energy and tool-wear penalties. The resulting traversal values are used as a baseline representative of the geometric efficiency achieved by the CQM under the DRP constraints.

As an external reference, and solely for comparison purposes, we also solve a classical distance-only TSP using Google OR-Tools on the same set of hole coordinates. This solution represents the shortest purely geometric tour without any manufacturing constraints.  

In Table~\ref{tab:objective_quality_split}, the \textit{Traversal baseline} column shows the mean traversal of the best CQM solutions and, in parentheses, its percentage gap with respect to the OR-Tools TSP reference:
\begin{equation}
    \text{GeometricGap}(\%) \;=\;
    \frac{\overline{\text{Traversal}}_{\mathrm{CQM}} - \overline{\text{Traversal}}_{\mathrm{TSP}}}
         {\overline{\text{Traversal}}_{\mathrm{TSP}}}\times 100.
\end{equation}
This gap isolates the geometric overhead induced by routing under the DRP
constraints relative to the shortest unconstrained geometric tour.

\paragraph{Penalty gap (blocks and tool-wear)} To quantify the energy surcharge introduced by block enforcement and wear accumulation with respect to pure displacement, we consider the difference between the mean best total energy and the mean traversal:
\begin{equation}
    \text{PenaltyGap} \;=\;
    \overline{\text{TotalEnergy}}_{\mathrm{CQM}} \;-\; \overline{\text{Traversal}}_{\mathrm{CQM}},
\end{equation}
This measure disentangles displacement-related efficiency from physics-informed penalties, offering an industrially interpretable indicator.

\paragraph{Discussion} Across instances, longer time limits (i.e., solver times of 15s and 20s) typically reduce both the mean best total energy and the geometric gap reported alongside the traversal values. This indicates an improved reconciliation between geometric routing efficiency and the physics‑informed terms. Also, the \emph{Penalty gap} quantifies the energetic surcharge attributable to block enforcement and tool-wear accumulation; we observe that this surcharge remains bounded and tends to decrease as time increases, suggesting that the hybrid workflow exploits the additional budget to shape more favourable block patterns while controlling wear growth.  These joint trends confirm that higher time budgets allow the solver to refine displacement, better exploit material grouping to reduce tool changes, and accommodate wear penalties without compromising feasibility.

\subsection{Structural analysis of material blocks}
\label{subsec:structural_quality}

Since the DRP enforces material grouping through block-related constraints, indicators such as the number of blocks, their average length, and the resulting number of tool changes directly affect manufacturability and operational efficiency. To this end, we consider the solutions obtained with a 15-second time limit, which provides a representative trade-off between solution quality and computational cost. For each instance, we perform 10 independent runs and compute the mean number of material blocks, average block length, and tool changes from the best feasible solution of each run. These values are compared against a naive non-grouped baseline.

\begin{table}[t]
\centering
\caption{Structural indicators of material grouping under the 15-second configuration.
Values are means over $K=10$ runs; in parentheses, comparison with a non-grouped baseline.}
\label{tab:structural_blocks}
\setlength{\tabcolsep}{5pt}
\resizebox{0.65\columnwidth}{!}{
\begin{tabular}{lccc}
\toprule
\textbf{Instance} &
\textbf{\# Blocks} &
\textbf{Avg. block length} &
\textbf{\# Tool changes} \\
\midrule
DRP\_10 & 2.0 (4.7) & 5.00 (2.30) & 1.0 (3.7) \\
DRP\_15 & 2.0 (9.4) & 7.50 (1.73) & 1.0 (8.4) \\
DRP\_20 & 3.1 (11.0) & 6.50 (1.82) & 2.1 (10.0) \\
DRP\_25 & 4.4 (17.8) & 5.92 (1.41) & 3.4 (16.8) \\
DRP\_30 & 4.9 (21.4) & 6.30 (1.41) & 3.9 (20.4) \\
DRP\_35 & 4.9 (21.9) & 7.41 (1.60) & 3.9 (20.9) \\
DRP\_40 & 5.1 (23.3) & 7.93 (1.73) & 4.1 (22.3) \\
DRP\_45 & 5.8 (23.5) & 8.08 (1.92) & 4.8 (22.5) \\
DRP\_50 & 6.4 (32.0) & 8.01 (1.58) & 5.4 (31.0) \\
DRP\_55 & 6.8 (38.1) & 9.06 (1.46) & 5.8 (37.1) \\
\bottomrule
\end{tabular}}
\end{table}

The results show that the CQM-based solutions consistently yield a markedly lower number of tool changes than the naive baseline. This reduction is a direct consequence of the block-continuation and block-start constraints, which incentivise longer homogeneous segments and prevent premature material switching. In addition, the block lengths observed across instances remain stable and coherent: they are long enough to minimise unnecessary tool exchanges but short enough to remain compatible with the geometric shape of the structure and the imposed origin-time ordering.

Overall, this structural analysis highlights the practical value of the CQM formulation. By integrating block-related constraints directly into the model, the solver generates drilling trajectories that naturally respect manufacturability principles-reducing tool-change overhead, concentrating wear progression within controlled segments, and preserving the balance between material grouping and geometric efficiency.

\section{Conclusions and Future Work} \label{sec:conc}

This work presented a hybrid quantum-classical approach for the DRP in multi-material structures by formulating it as a CQM. Experimental results demonstrate that the method consistently generates feasible, energy-efficient drilling trajectories that satisfy operational constraints, reduce tool changes, and control wear progression. Future work will focus on integrating more complex machining constraints moving further toward quantum-enabled CAM frameworks for advanced manufacturing



\bibliographystyle{IEEEtran}
\bibliography{IEEEexample}
\end{document}